\documentclass[11pt,a4paper]{article}
\usepackage[hyperref]{emnlp-ijcnlp-2019}
\usepackage{times}
\usepackage{latexsym}
\usepackage{amsmath}

\usepackage{url}

\aclfinalcopy 

\title{Multi-Granularity Representations of Dialog}

\author{Shikib Mehri and Maxine Eskenazi \\
  Language Technologies Institute, Carnegie Mellon University \\
  {\tt \{amehri,max+\}@cs.cmu.edu}} 
\date{}

\begin{document}
\maketitle
\begin{abstract}
Neural models of dialog rely on generalized latent representations of language. This paper introduces a novel training procedure which explicitly learns multiple representations of language at several levels of granularity. The multi-granularity training algorithm modifies the mechanism by which negative candidate responses are sampled in order to control the granularity of learned latent representations. Strong performance gains are observed on the next utterance retrieval task using both the MultiWOZ dataset and the Ubuntu dialog corpus. Analysis significantly demonstrates that multiple granularities of representation are being learned, and that multi-granularity training facilitates better transfer to downstream tasks.

\end{abstract}

\section{Introduction}
Producing generalized representations of language is a well-studied problem in natural language processing (NLP) \citep{montague1973proper,davidson2012semantics}. Neural models typically encode an input into a latent vector, which is then used by upper layers. As such, improving the quality or generality of the learned representations will typically improve performance on the final task due to the increased representative power of the model. 

Constructing meaningful representations of dialog is challenging. To effectively represent the dialog context, a latent dialog representation must contain the information necessary to (1) estimate a belief state over user goals \citep{williams2013dialog}, (2) track entity mentions \citep{zhao2017generative}, (3) resolve anaphora co-references \citep{mitkov2014anaphora}, (4) model the communicative purpose of an utterance \citep{core1997coding} and (5) resolve ambiguity in natural language. A large focus area of dialog research is the development of neural architectures which learn effective representations of the input \citep{zhou2016multi,wu2016sequential,zhou2018multi}. With the goal of training a model for next utterance retrieval, \citet{zhou2018multi} use a deep self-attention network to produce a representation of each utterance within a dialog and follow it with an attention between utterances and 3-D convolutional layers. 

Recent work has explored the use of large-scale self-supervised pre-training on very large corpora \citep{kiros2015skip,peters2018deep,devlin2018bert,radford2018improving} as a means of improving natural language representations. These pre-trained models have yielded state-of-the-art results on several downstream NLP tasks \citep{wang2018glue}: text classification, natural language inference, and question answering. Though such methods have proven useful across several downstream tasks \citep{wang2018glue}, using them for dialog requires expensive fine-tuning of the complex models \citep{dinan2019second,alberti2019bert}. The need for this fine-tuning is due to the pre-training procedure. First, the domain and style of dialog corpora differ significantly from the majority of the data used during pre-training. This necessitates fine-tuning in order to adapt the representations to more varied input. Second, the pre-trained representations, which are all obtained through various language modelling objectives, do not necessarily capture properties of dialog at several levels of granularity (e.g., belief state, entities, co-references, high-level user goals). 

Though large-scale pre-training improves the strength and generality of latent representations, this effect is minimized when transferring to dialog tasks or out-of-domain data. To this end, this paper explores an alternate mechanism of learning strong and general representations for the task of next utterance retrieval \citep{lowe2015ubuntu}. We propose Multi-Granularity Training (MGT), which simultaneously trains multiple levels of representation. It later combines these latent representations to obtain more general models of dialog. Different granularities of representation capture different properties of the input. For example, a high-granularity representation will capture specific words and entities mentioned in the dialog context. A low-granularity representation will instead capture more abstract properties of the dialog, such as the domain of the conversation or the high-level user goal. MGT combines representations at several levels of granularity, resulting in stronger and more general representations of dialog. The strength of representations is a consequence of learning the dedicated representations at each level of granularity. The generality results from learning several diverse representations across multiple granularities, thereby encompassing a wider amount of information. Since the representations are learned on dialog data and for the final task, this method does not suffer from the aforementioned shortcomings of pre-training. 

The specific MGT procedure is motivated by the fact that observing different negative examples during training results in different representations. A model trained to select the correct response out of a set of lexically similar candidates will likely learn fine-grained representations of each word in an effort to identify minute differences between the candidates. On the other hand, a model trained to select a response from a set of topically diverse candidates will likely learn broader and more abstract representations of each utterance. Typically, negative examples are randomly sampled which results in learned representations that fit the average training example. MGT relies on an algorithm for controlled sampling of negative candidate responses, which allows for the construction of multiple training sets in order to learn multiple levels of granularity. 

MGT is agnostic to the underlying model architecture. Though the majority of experiments in this paper are carried out with a dual encoder \citep{lowe2015ubuntu} as the base model, MGT is also applied on top of Deep Attention Matching networks \citep{zhou2018multi} and obtains strong performance gains.

MGT is evaluated using the MultiWOZ dataset \citep{budzianowski2018multiwoz} and the Ubuntu dialog corpus \citep{lowe2015ubuntu} to train models for next utterance retrieval. Results show that MGT obtains better performance than ensembling \citep{perrone1992networks} multiple baseline models. At the same time, it also serves as a better downstream representation of dialogs. The contributions of this paper are: (1) a training procedure which learns multiple granularities of latent representations for a task, (2) improved performance on next utterance retrieval across two diverse datasets, (3) an analysis significantly demonstrating that multiple granularities of representation have indeed been learned.

\section{Related Work}

This section discusses two areas of related work: language representations and the next utterance retrieval task.  

\subsection{Language Representations}

Recent work has focused on improving latent representations of language through the use of large-scale self-supervised pre-training on very large corpora. \citet{kiros2015skip} trains a sequence-to-sequence model \citep{sutskever2014sequence} to predict the surrounding sentences, and uses the final encoder hidden state as a generic sentence representation. ELMo \citep{peters2018deep} trained a bi-directional language model on a large corpus in order to obtain strong contextual representations of words. OpenAI's GPT \citep{radford2018improving} produces latent representations of language by training a large transformer \citep{vaswani2017attention} with a language modelling objective. \citet{devlin2018bert} further improves on this line of research by introducing the masked language modelling objective and a multi-tasking pre-training loss. Each of these methods has obtained state-of-the-art results on the GLUE benchmark \citep{wang2018glue}, suggesting that they are strong and general representations of language.

These pre-trained representations of language have been applied to numerous tasks. Of particular interest are applications of these representations to dialog tasks. As part of the 2nd ConvAI challenge \citep{dinan2019second}, the best performing models on both human and automated evaluations  \citep{wolf2019transfertransfo} were fine-tuned versions of OpenAI's GPT \citep{radford2018improving}. Despite strong performance gains, transferring OpenAI's GPT required fine-tuning the full model because the dialog data was in a different domain and required different information to be contained in the representations. Recently, \citet{mehri2019pretraining} introduce several dialog specific pre-training objectives that obtain strong performance gains across multiple downstream dialog tasks.

\subsection{Next Utterance Retrieval}

 \citet{lowe2015ubuntu} construct Ubuntu, the largest retrieval corpus for dialog, and present the dual encoder architecture as a baseline architecture. \citet{kadlec2015improved} present several strong baseline architectures for this dataset. \citet{zhou2016multi} present the Multiview architecture which, with the aim of constructing broader representation, learns both word-level representations and utterance-level representations. Sequential Matching Networks (SMN) \citep{wu2016sequential} represent each utterance in the dialog context and construct segment-segment matching matrices between the response and each utterance in the context. Deep Attention Matching (DAM) \citep{zhou2018multi} uses deep transformers \citep{vaswani2017attention} to construct representations of each utterance in a dialog context, followed by cross-attention and convolutional layers. 

Previous work on next utterance retrieval has proposed architectural modifications in an effort to improve the representative powers of the models. This paper presents a training algorithm applicable to any neural architecture, which explicitly forces the model to learn different granularities of representation. 

\section{Methods}

This section describes three methods used for next utterance retrieval: a strong baseline dual encoder architecture, an ensemble of dual encoders, and an ensemble of dual encoders with multi-granularity training.

\subsection{Dual Encoder}

Given a dialog context, next utterance retrieval selects the correct response from a set of $k$ candidates. The retrieval baselines presented by \citet{kadlec2015improved} first encode the dialog context and a candidate response. Then they use the product of the latent representations to output a probability. This baseline architecture consists of two encoders, one to encode the context and one for the response. 

Previous approaches using Ubuntu were trained for binary prediction (i.e., predict the probability of a particular response), and used during testing to select from a candidate set. To mitigate the discrepancy between training and testing, our baseline is \textit{trained} to select the correct response from a candidate set. Since the Ubuntu training set consists of 0/1 labels, the training set was modified by considering only the positive-labeled examples, and uniformly sampling $k-1$ negative candidates. 

Let $c_{1,\dots,N}$ denote the words of the dialog context, $r^i_{1,\dots,M_i}$ denote the words of the $i$-th candidate response and $r_{gt}$ denote the ground-truth response. Given $f_c$, the LSTM encoder of the context, and $f_r$, the LSTM encoder of the candidate responses, the forward propagation of the dual encoder is described by:

\begin{align}
    \mathbf{c} &= f_{\text{c}}(c_i) \quad i \in [1, N]\\
    \mathbf{r_i} &= f_{\text{r}}(r^i_j) \quad j \in [1, M_i]\\
    \mathbf{r_{gt}} &= f_{\text{r}}(r^{gt}_j) \quad j \in [1,M_{gt}]\\  
    \alpha_{gt} &= \mathbf{c}^T \mathbf{r_{gt}}\\
    \alpha_{i} &= \mathbf{c}^T \mathbf{r_{i}}
\end{align}

The final loss function is:

\begin{align}
    \mathcal{L} &= -\log p(r^i_{1,\dots,M_i}|c_{1,\dots,N}) \\ \nonumber
            &= -\log \left( \frac{\exp(\alpha_{gt})}{\exp(\alpha_{gt})+\sum_{j=1}^K \exp(\alpha_{j})} \right)
\end{align}

\subsection{Ensemble of Dual Encoders}

Ensembling multiple models \citep{perrone1992networks} has been empirically shown to improve performance, since it maintains a low model bias while significantly reducing the model variance. In ensembling, multiple models are trained and their predictions are averaged during inference. Specifically, if $\alpha^l$ denotes the output of model $l \in [1,L]$, the output probability is defined as:

\begin{align}
\label{eq:ensemble}
    p(r^i_{1,\dots,M_i}|c_{1,\dots,N}) = \frac{1}{L} \sum_{l=1}^{L} \frac{exp(\alpha^l_i)}{\sum_{j=1}^K exp(\alpha^j_i)}
\end{align}

Since ensembling reduces the model variance while maintaining low bias, it is most effective when the models are diverse and each model excels at a particular type of input. In typical ensemble training, the different models are either obtained through different random initializations or at different checkpoints from the same training run. In such an approach, there is no mechanism which explicitly enforces diversity between the models.

\subsection{Multi-Granularity Training}

During baseline model training, the negative response candidates were uniformly sampled from $R$, the set of all responses in the training set. MGT is proposed in an effort to explicitly model different granularities of representation through a controlled method of sampling negative candidates. 

Consider a training corpus consisting of a set of dialog contexts and ground-truth responses, $T = (C, R^{gt})$. In the baseline training, $k-1$ negative response candidates are uniformly sampled from the set of all responses, $R$:

\begin{align}
    T_i = (C_i, R^{gt}_i, [N_{i,1}, N_{i,2}, \dots, N_{i,k-1}]) \\ \nonumber
    \forall j \in [1,k-1] ~~~~ N_{i,j} \sim \text{Uniform}(R)
\end{align}

MGT is motivated by the idea that observing different types of negative candidate response sets will result in different representations. Negative candidates which are lexically similar to the ground truth response should result in models that carefully consider each word in order to produce fine-grained representations and identify minute differences between candidate responses. On the other hand, very semantically distant candidate responses should result in very broad and abstract representations of language. While there may be many methods of sampling negative responses to influence what the model learns, this paper focuses on using the semantic similarity of the candidate responses as a means of controlling the granularity of learned representations. 

Given the LSTM response encoder, $f_r$, the measure of semantic similarity is defined as:

\begin{align}
    \mathbf{r_i} &= f_r(R_{i,j}) \quad j \in [1,M_i] \\
    \mathbf{r_k} &= f_r(R_{k,j}) \quad j \in [1,M_k] \\
    d(R_i, R_k) &= \frac {\mathbf{r_i}^T \mathbf{r_k}}{||\mathbf{r_i}|| \cdot ||\mathbf{r_k}||}
\end{align}

This approach relies on a cosine-similarity as a measure of semantic distance between dialog utterances. While not a perfect measure, for the purposes of the MGT algorithm it appears to be a sufficient measure. Since the training algorithm groups together similarly distant negative candidates, it is robust to noise in the measure of semantic distance. Future work may explore whether a better distance measure improves the MGT algorithm.

A distance matrix $D$ is constructed between all of the responses in $R$, such that $D_{i,j} = d(R_i, R_j)$. The objective of MGT is to train $L$ models at $L$ different levels of granularity. For a particular response $R_i$, rather than sampling negative candidates from the entire set of $R$, the set of responses $R$ is split into $L$ segments based on distance from $R_i$. Define a function ${\tt b}(D_i, l)$ which considers a list of distances and returns the maximum distance in the $l$-th segment of a total of $L$ segments. This is equivalent to sorting $D_i$ and taking the $\left(|R|\times \frac{l}{L}\right)$-th value. 

The distance matrix, $D$, is used to segment the set of potential negative candidates, $R$, for each training example $(C_i, R^{gt}_i)$, into $L$ buckets: $P^1_i,\dots,P^L_i$. Given the definition of segmentation provided above, $P^1_i$ will consist of responses that are strictly closer (as defined by $d$) to $R_i$ than the responses in $P^2_i$. When training the $l$-th model at the $l$-th level of granularity, the negative responses for $R_i$ are sampled from $P^l_i$ rather than $R$. $P^l_i$ is constructed using ${\tt b}(D_i, l)$, which was defined to return the maximum value in the $l$-th segment.

This method is used to construct $L$ different training corpora, $T^1,\dots,T^L$. A particular $T^l$ is constructed as follows:

\begin{align}
    &T^l_i = (C_i, R^{gt}_i, [N^l_{i,1}, \dots, N^l_{i,k-1}])  \\  \nonumber
    &P^l_i = \{ r \in R ~|~ d(R_i, r) \in ({\tt b}(D_i, l-1), {\tt b}(D_i, l) \}\\ \nonumber
    &\forall j \in [1,k-1] ~~~~ N^l_{i,j} \sim \text{Uniform}(P^l_i)
\end{align}

After the $L$ different training corpora, $L$ different models are trained. Models trained on closer candidate sets should learn more granular representations while models trained on more distant candidate sets should learn more abstract representations of dialog. Upon obtaining $L$ different models, the output probability is produced by the ensembling method described in Equation \ref{eq:ensemble}.

\section{Experiments}

This section describes the datasets and presents experimental procedures aimed at evaluating the different approaches to next utterance retrieval. 

\subsection{Datasets}

Two retrieval corpora, MultiWOZ \citep{budzianowski2018multiwoz} and Ubuntu \citep{lowe2015ubuntu} were used. MultiWOZ contains task-oriented conversations between a tourist and a Wizard-of-Oz, while Ubuntu contains both open-domain and technical dialog snippets collected from Internet Relay Chat (IRC). The diversity of these two datasets provides insight into the general applicability of MGT.

\subsubsection{MultiWOZ}

The MultiWOZ dataset \cite{budzianowski2018multiwoz} was converted into a retrieval corpus. MultiWOZ contains 8422 dialogs for training, 1000 for validation and 1000 for testing. There are 20 candidate responses for each dialog context.

\subsubsection{Ubuntu Dialog Corpus}

The original Ubuntu corpus \citep{lowe2015ubuntu} has 1,000,000 training examples. Typical interactions include individuals asking for technical assistance in a conversational manner. The subject of conversation is not explicitly bounded and may be any topic. As described in Section 3.1, the training corpus is modified in order to train as a retrieval task rather than as a binary prediction task. Negative training examples (500,127) are filtered out. The size of the new training dataset is 499,873. There are a total of 10 candidate responses for each context. The validation and test sets remain unchanged, with 19,561 validation examples and 18,921 test examples.

\subsection{Experimental Setup}

Unless otherwise specified, the size of ensembles and the number of models in MGT is $L=5$. For MGT, the highest performing checkpoint at each granularity is selected using the validation score. For the ensemble method, the top performing checkpoints are selected from a single run.

\subsubsection{MultiWOZ Setup}

Two distinct encoders are trained, one to encode the dialog context and the other for the candidate responses. Each encoder is a single layer, uni-directional LSTM with an embedding dimension of 50 and a hidden size of 150. These hidden sizes match the best performing hyperparameters identified by \citet{budzianowski2018multiwoz}. The Adam optimizer \citep{kingma2014adam} with a learning rate of 0.005 is used to train the model for 20 epochs. The vocabulary is 1261 words, the batch size is 32, and gradients are clipped to 5.0. A checkpoint is saved after each epoch, and the best checkpoint is selected using performance on the validation set.

\subsubsection{Ubuntu Setup}

Each encoder is a single layer, uni-directional LSTM with an embedding dimension of 300 and a hidden size of 150. The Adam optimizer \citep{kingma2014adam} with a learning rate of 0.005 is used to train the model for 20 epochs. The vocabulary is 10002 words, the batch size is 128, and gradients are clipped to 5.0. Only the last 160 words of each dialog context are used. The word embeddings are initialized with pre-trained GloVe embeddings \citep{pennington2014glove}. A checkpoint is saved after each epoch, and the best checkpoint is selected using performance on the validation set.

\subsubsection{Deep Attention Matching Experiment}

MGT is a training procedure which is agnostic to the underlying model architecture. Though the majority of the experiments presented in this paper use the dual encoder architecture \citep{lowe2015ubuntu}, MGT is applied on top of the state-of-the-art architecture for Ubuntu: the Deep Attention Matching Network (DAM) \citep{zhou2018multi}. When applying MGT to DAM, the sampling of negative candidates is done using the baseline dual encoder architecture.

\subsection{Multi-Granularity Training Example}

\begin{table*}[]
    \centering
    \begin{tabular}{l}
    \hline \hline
\textbf{Dialog Context:}\\
\texttt{USER:} i need an expensive place to dine in the center of town . \\
\texttt{SYS:} what type of food are you wanting at this restaurant ?  \\
\texttt{USER:} type of food does not matter to me . \\
\hline
\textbf{Ground-Truth Response:} \\
\texttt{SYS:} yes there are 33 such restaurant -s . would  you like me to book 1 for you ? \\
\hline
\textbf{Closest Neg. Cand. (for Highest Granularity Model):} \\
\texttt{SYS:} i have 5 indian restaurant -s in the west . if i may , might i suggest the rice boat . i can reserve\\ a table for you if you would like . \\
\hline
\textbf{2nd Closest Neg. Cand. (for 2nd Highest Granularity Model):} \\
\texttt{SYS:} how about the avalon ? \\
\hline
\textbf{3rd Closest Neg. Cand. (for 3nd Highest Granularity Model):} \\
\texttt{SYS:} i would be happy to book that for you .are you travelling alone ? \\
\hline
\textbf{2nd Furthest Neg. Cand. (for 2nd Lowest Granularity Model):} \\
\texttt{SYS:} i apologize , there s been a misunderstanding . before we can book your taxi , we need to know \\which college . there are 18 colleges . what area are you looking for ? \\\hline
\textbf{Furthest Granularity Neg. Cand (for Lowest Granularity Model):} \\
\texttt{SYS:} i need to know when you want to leave and  arrival time in order for me to order you a taxi . \\

\hline 

\hline \hline
\end{tabular}
\caption{An example dialog context from the training set. Along with the ground-truth response, negative candidates sampled at five different levels of semantic distance are shown. The retrieval models are trained to differentiate between the ground-truth response and the different negative candidates.}
\label{mwoz_example}
\end{table*}

Table \ref{mwoz_example} shows an example dialog context, ground-truth response and the negative candidate responses sampled at several levels of cosine distance, as per Equation 11. These negative candidate responses are retrieved by MGT's sampling algorithm, and are used to train multiple models at different levels of granularity.

The negative candidates for the higher granularity models are much closer to the ground-truth response than the candidates for the lower granularity models. As such, models trained at higher granularities will learn more granular representations of dialog in order to identify minute differences between responses. Examples for lower granularity models are more distant in meaning, and models may instead learn to represent higher-level attributes of the dialog.

\section{Results}

Multi-granularity training is proposed in order to learn strong and general latent representations of dialog. To evaluate the strength and generality of the learned representations, experiments are conducted to evaluate three different properties of MGT: (1) improved performance on the task of next utterance retrieval, (2) explicit modelling of different granularities, and (3) improved generality and transferability to other dialog tasks.


\subsection{Next Utterance Retrieval}

Next utterance retrieval is reliant on latent representations of dialog. Several experiments are conducted to evaluate whether MGT improves the representative power of models and results in better performance on the task of next utterance retrieval. MGT is expected to outperform standard ensembling, since MGT explicitly models multiple granularities and trains more diverse models. The performance of MGT is evaluated using both MultiWOZ \citep{budzianowski2018multiwoz} and Ubuntu \citep{lowe2015ubuntu}. Experiments are conducted using two different underlying architectures, a dual encoder baseline \citep{lowe2015ubuntu} and a Deep Attention Matching network \citep{zhou2018multi}.

\subsubsection{MultiWOZ}

Performance on the MultiWOZ retrieval task is evaluated with mean reciprocal rank (MRR), and Hits@1 (H@1). Mean reciprocal rank is defined as follows:

\begin{equation}
  MRR = \frac{1}{N} \sum_{i=1}^N \frac{1}{rank_i}
\end{equation}

Hits@1 is equivalent to accuracy. It measures how often the ground-truth response is selected from the $K=20$ candidates. 

The results in Table \ref{mwoz_results} demonstrate the strong performance gains obtained with MGT. With $L=5$ granularities, MGT outperforms a similarly sized ensemble of dual encoders. These results demonstrate that explicitly enforcing the policy that makes models learn multiple granularities of representation improves the representative power and performance on next utterance retrieval.

\begin{table}[]
\centering
\begin{tabular}{|l|l|l|}
\hline
\textbf{Model Name}            & \textbf{MRR} & \textbf{Hits@1} \\ \hline
Dual Encoder                   & 79.55        & 66.13\%         \\ 
Ensemble  (5)   & 81.53        & 69.47\%         \\ 
Multi-Granularity (5) & \textbf{82.74}        & \textbf{72.18}\%         \\ \hline
\end{tabular}
\caption{Performance on MultiWOZ. MGT is compared to a baseline dual encoder, and an ensemble of dual encoders with an identical number of parameters. All bold-face results are statistically significant to $p < 0.01$.}
\label{mwoz_results}
\end{table}

\begin{table*}[]
\centering
\begin{tabular}{|l|c|c|c|}
\hline
\textbf{Model Name}            & \multicolumn{1}{l|}{\textbf{MRR}} & \multicolumn{1}{l|}{\textbf{$\textbf{R}_{10}@1$}} & \multicolumn{1}{l|}{\textbf{$\textbf{R}_{2}@1$}} \\ \hline

\multicolumn{4}{|c|}{\textbf{Previous Research}} \\ \hline

Dual Encoder \citep{lowe2015ubuntu}                  & -                                 & 63.8                               & 90.1                                \\                   
MV-LSTM \citep{pang2016text}                          & -                                 & 65.3                               & 90.6                                \\ 
Match-LSTM \citep{wang2016machine}                     &                                   & 65.3                               & 90.4                                \\ 
Multiview \citep{zhou2016multi}                      & -                                 & 66.2                               & 90.8                                \\ 
DL2R \citep{yan2016learning}                           & -                                 & 62.6                               & 89.9                                \\ 
SMN \citep{wu2016sequential}                           & -                                 & 72.6                               & 92.6                                \\ 
DAM \citep{zhou2018multi}                            & -                                 & \textbf{76.7}                              & \textbf{93.8}                                \\ \hline
\multicolumn{4}{|c|}{\textbf{Dual Encoder Experiments}} \\ \hline
Dual Encoder \citep{lowe2015ubuntu}                   & 76.84                             & 63.6                              & 90.9                               \\ 
Ensemble  (5)                 & 78.91                             & 66.9                              & 91.7                               \\ 
Multi-Granularity (5)        & \textbf{80.10}                            & \textbf{68.7}                              & \textbf{91.9}                              \\ \hline
\multicolumn{4}{|c|}{\textbf{Deep Attention Matching Experiments}} \\ \hline
DAM \citep{zhou2018multi} (re-trained)                 & 83.74                             & 74.54                              & 93.08                               \\ 
Ensemble  (5)                 & 84.03                             & 74.95                              & 93.27                               \\ 
Multi-Granularity (5)        & \textbf{84.26}                            & \textbf{75.30}                              & \textbf{93.45}                              \\ \hline

\end{tabular}
\caption{Results for next utterance retrieval on the Ubuntu dialog corpus. This table shows previous work, and experimental results with two underlying architectures: a dual encoder model and Deep Attention Matching networks.  The results shown in the DAM experiments section are performed with the open-sourced implementation of \citet{zhou2018multi}, which obtains slightly worse performance than they report. All bold-face results are statistically significant to $p < 0.01$.}
\label{ubuntu_results}
\end{table*}

\subsubsection{Ubuntu}

Previous research used several variations of the $R_N@k$ metric to evaluate retrieval performance on the Ubuntu dialog dataset. $R_N@k$ refers to the percentage of the time that the ground truth response was within the top-$k$ predictions for a candidate set size of $N$ utterances. $R_{10}@1$ on Ubuntu is equivalent to Hits@1 and accuracy. In addition to MRR, we report $R_{10}@1$ and $R_2@1$, top-1 accuracy with a candidate set size of 10 and 2, respectively.

MGT is applied on top of the dual encoder baseline \citep{lowe2015ubuntu} and Deep Attention Matching networks \citep{zhou2018multi}. The results shown in Table \ref{ubuntu_results} show the performance of MGT using two different underlying architectures, as well as previous work. Across both base architectures, MGT outperforms ensembling. The primary difference between these two methods is that MGT explicitly ensures that several granularities of representation are learned. As such, these results reaffirm the hypothesis that learning multiple granularities of representation leads to more diverse models, and more general representations of dialog.  

Even with the dual encoder as the underlying model, MGT outperforms all previous work except for Sequential Matching Networks (SMN) \citep{wu2016sequential} and Deep Attention Matching networks (DAM) \citep{zhou2018multi}. The Deep Attention Matching experiment performs MGT using DAM\footnote{It should be noted that the open-source implementation provided by \citet{zhou2018multi} was used, however performance was slightly lower than the results they reported. We speculate that given a DAM implementation that matches their reported results, MGT would obtain a similarly-sized improvement (+0.76 R@1).} as the underlying architecture. MGT has good performance improvement on top of DAM, roughly double the improvement obtained by ensembling. This suggests that MGT can be used as a general purpose training algorithm which learns multiple-granularities of representation and thereby produces stronger and more general models.




\subsection{Explicit Granularity Modelling}

Multi-granularity training learns multiple granularities of representation. However, strong performance on next utterance retrieval, does not necessarily prove that several granularities are explicitly modelled. 
To analyze whether the models operate at different levels of granularity, the content of the representations must be considered. The $L=5$ trained models, each at a different granularity, have their weights frozen. These frozen models are then used to obtain a latent representation of all the dialog contexts in MultiWOZ. A linear layer is then trained on top of these representations for a downstream task. During this training, only the weights of the linear layer are updated. This evaluates the information contained in these learned representations.

Two different downstream tasks are considered; bag-of-words prediction and dialog act prediction. Bag-of-words prediction is the task of predicting a binary vector corresponding to the words present in the last utterance of the dialog context. This task requires very granular representations of language, and therefore the models trained to capture high granularity representations should have the highest performance. Dialog act prediction is the task of predicting the set of dialog acts for the next system response. This is a high-level task that requires abstract representations of language, therefore the models with the lowest granularity should do well. 

The results in Table \ref{granularity} confirm the hypothesis that MGT results in models that learn different granularities of representation. It is clear that higher granularity models better capture the information necessary for the bag-of-words task, while higher abstraction (lower granularity) models better capture information for dialog act prediction.

\begin{table}[]
\centering
\begin{tabular}{|l|c|c|}
\hline
\textbf{Model Name}             & \textbf{BoW (F-1)} & \textbf{DA (F-1)} \\ \hline
Highest Abstraction             & 57.00        & \textbf{19.24}         \\ 
2nd Highest Abs.         & 57.69        & 19.14         \\ 
Medium                          & 58.49        & 18.31         \\ 
2nd Highest Gran.         & 58.38        & 16.88         \\ 
Highest Granularity             & \textbf{59.43}        & 15.46         \\  \hline
\end{tabular}
\caption{Results of the granularity analysis experiment. $L=5$ models trained to capture different granularities of representation. All bold-face results are statistically significant to $p < 0.01$.}
\label{granularity}
\end{table}

\subsection{Generalizability and Task Transfer}

One motivation of MGT is to improve the generality of representation, and facilitate easy transfer to various tasks. Truly general representations of language would require no fine-tuning of the model, and we would only need to learn a linear layer in order to extract the relevant information from the representation. Bag-of-words prediction and dialog act prediction are again used to evaluate the ability of MGT to transfer without any fine-tuning.

\begin{table}[]
\centering
\begin{tabular}{|l|c|c|}
\hline
\textbf{Model Name}             & \textbf{BoW (F-1)} & \textbf{DA (F-1)} \\ \hline
Dual Encoder             & 60.13        & 19.09         \\ 
Ensemble (5)         & 64.11        & 22.39     \\ 
Multi-Granularity (5)                          & \textbf{67.51}        & \textbf{22.85}         \\ \hline
Fine-tuned  & 90.33 & 28.75 \\ \hline
\end{tabular}
\caption{Experimental results demonstrating performance on two downstream tasks, without any fine-tuning of the latent representations. All bold-face results are statistically significant to $p < 0.01$.}
\label{finetune}
\end{table}

The results shown in Table \ref{finetune} demonstrate that MGT results in more general representations of language, thereby facilitating better transfer. However, there is room for improvement when comparing to models fine-tuned on the downstream task. This suggests that additional measures can be taken to improve the representative power of these models. 


The results in Table \ref{finetune2} demonstrate that MGT learns general representations which effectively transfer to downstream tasks, especially more difficult tasks such as dialog act prediction. Fine-tuning the latent representations learned by MGT, results in improved performance on dialog act prediction.


\begin{table}[]
\centering
\begin{tabular}{|l|c|c|}
\hline
\textbf{Model Name}              & \textbf{DA (F-1)} \\ \hline
Random Init                     & 28.75        \\ 
Dual Encoder                     & 32.63       \\ 
Ensemble (5)                 & 31.71     \\ 
Multi-Granularity (5)                                  & \textbf{33.46}         \\ \hline

\end{tabular}
\caption{Experimental results demonstrating performance on the downstream task of dialog act prediction, when the model is fine-tuned on all available data. All bold-face results are statistically significant to $p < 0.01$.}
\label{finetune2}
\end{table}

\section{Conclusions and Future Work}

This paper presents multi-granularity training (MGT), a mechanism for learning strong and general representations for next utterance retrieval. Through the use of a sampling algorithm to select negative candidate responses, multiple granularities of representation are learned during training. Strong performance gains are observed on the task of next utterance retrieval on both MultiWOZ and Ubuntu. Experiments show that MGT is a generally applicable training procedure which can be applied to multiple underlying model architectures. Quantitative analytic experiments demonstrate that multiple granularities of representation are in fact being learned, and that MGT facilitates better transfer to downstream tasks both with and without fine-tuning.

There are several avenues for future work. First, this method is general and broadly applicable, which suggests that it may improve performance on other tasks and domains. A particularly interesting application would be to generalize this method to language generation tasks. Second, a useful improvement on top of MGT would be a more sophisticated method of combining the multiple granularities of representations.  Third, while this paper focuses on capturing multiple representations at different levels of granularity, it would be interesting to generalize MGT to learning multiple representations along several different axes (e.g., domains, styles, intents, etc.).

\bibliography{emnlp-ijcnlp-2019}
\bibliographystyle{acl_natbib}

\end{document}